\documentclass{article} 
\usepackage{arxiv}

\usepackage{amsmath,amsfonts,bm}

\def\eqref#1{equation~\ref{#1}}

\def\1{\bm{1}}

\DeclareMathAlphabet{\mathsfit}{\encodingdefault}{\sfdefault}{m}{sl}
\SetMathAlphabet{\mathsfit}{bold}{\encodingdefault}{\sfdefault}{bx}{n}

\usepackage[utf8]{inputenc} 
\usepackage[T1]{fontenc}    
\usepackage{hyperref}       
\hypersetup{
    colorlinks=true,
    citecolor=blue
}
\usepackage{url}            
\usepackage{booktabs}       
\usepackage{amsfonts}       
\usepackage{nicefrac}       
\usepackage{microtype}      
\usepackage{lipsum}		
\usepackage{graphicx}
\usepackage{natbib}
\usepackage{doi}

\usepackage[textsize=tiny]{todonotes}

\DeclareUnicodeCharacter{1EF3}{\`y}
\DeclareUnicodeCharacter{003BB}{TT}

\usepackage{xcolor}         
\usepackage[linesnumbered, ruled,vlined]{algorithm2e}
\usepackage{amsmath}
\usepackage{yhmath}
\usepackage{wrapfig}
\usepackage{threeparttable}
\usepackage{multicol}
\usepackage{multirow}
\usepackage{adjustbox}
\usepackage{subfigure}
\usepackage{setspace}
\usepackage{caption}
\usepackage{tikz}
\usetikzlibrary{tikzmark,calc,fit}
\usepackage{listings}
\usepackage{algorithmic}

\hypersetup{
colorlinks = true,
urlcolor=blue
}

\definecolor{codegreen}{rgb}{0,0.6,0}
\definecolor{codegray}{rgb}{0.5,0.5,0.5}
\definecolor{codepurple}{rgb}{0.58,0,0.82}
\definecolor{backcolour}{rgb}{0.95,0.95,0.92}
\lstdefinestyle{mystyle}{
    inputencoding=utf8,
    extendedchars=true,
    commentstyle=\color{codegreen},
    keywordstyle=\color{magenta},
    numberstyle=\tiny\color{codegray},
    stringstyle=\color{codepurple},
    basicstyle=\ttfamily\small,
    breakatwhitespace=false,         
    breaklines=true,                 
    captionpos=b,                    
    keepspaces=true,                 
    numbers=left,                    
    numbersep=5pt,                  
    showspaces=false,                
    showstringspaces=false,
    showtabs=false,                  
    tabsize=4,
    literate={λ}{{$\lambda$}}1
}
\SetCommentSty{mycommfont}
\definecolor{Gray}{rgb}{0.85, 0.85, 0.85}
\definecolor{lightGray}{rgb}{0.93, 0.93, 0.93}

\title{Federated Learning for Internet of Things: 

A Federated Learning Framework for On-device Anomaly Data Detection}

\author{Tuo Zhang*, Chaoyang He*, Tianhao Ma, Lei Gao, Mark Ma, Salman Avestimehr\\
Viterbi School of Engineering \\
University of Southern California \\
\texttt{\{tuozhang, chaoyang.he, tianhaom, leig, mzma, avestime\}@usc.edu}
}

\begin{document}
\maketitle
\begin{abstract}
Federated learning can be a promising solution for enabling IoT cybersecurity (i.e., anomaly detection in the IoT environment) while preserving data privacy and mitigating the high communication/storage overhead (e.g., high-frequency data from time-series sensors) of centralized over-the-cloud approaches. 
In this paper, to further push forward this direction with a comprehensive study in both algorithm and system design, we build \texttt{FedIoT} platform that contains \texttt{FedDetect} algorithm for on-device anomaly data detection and a system design for realistic evaluation of federated learning on IoT devices. Furthermore, the proposed \texttt{FedDetect} learning framework improves the performance by utilizing a local adaptive optimizer (e.g., Adam) and a cross-round learning rate scheduler. 
In a network of realistic IoT devices (Raspberry PI), we evaluate \texttt{FedIoT} platform and \texttt{FedDetect} algorithm in both model and system performance. Our results demonstrate the efficacy of federated learning in detecting a wider range of attack types occurred at multiple devices. The system efficiency analysis indicates that both end-to-end training time and memory cost are affordable and promising for resource-constrained IoT devices. The source code is publicly available at \url{https://github.com/FedML-AI/FedIoT}.
\end{abstract}

\section{Introduction}
Along with the faster Internet speed and more endpoints brought by the 5G, billions of IoT devices online will be deployed \cite{iot2025}. However, for the \textit{data anomaly detection task} (e.g., DDoS attack detection), the centralized over-the-cloud approach \cite{Meidan} may not fit this trend due to data privacy and extremely high communication/storage overhead (e.g., high-frequency data from time-series sensors) centralizing data from numerous IoT devices. As such, researchers attempt to address these challenges using federated learning (FL), which is a trending paradigm that can train a global or personalized model without centralizing data from edge devices \cite{kairouz2019advances}. DIoT \cite{diot} and IoTDefender \cite{fldef} employ FL for intrusion detection in IoT devices by collaboratively training isolated datasets for a global or even personalized model. 
\cite{9146846} further upgrades the model to a complex attention-based CNN-LSTM but mitigates the communication cost with Top-$k$ gradient compression. Beyond detecting the abnormal data, \cite{rey2021federated} even considers an adversarial setup where several malicious participants poison the federated model.


\begin{figure*}[h!]
\vskip 0.2in
\begin{center}
\centerline{\includegraphics[width=1.0\linewidth]{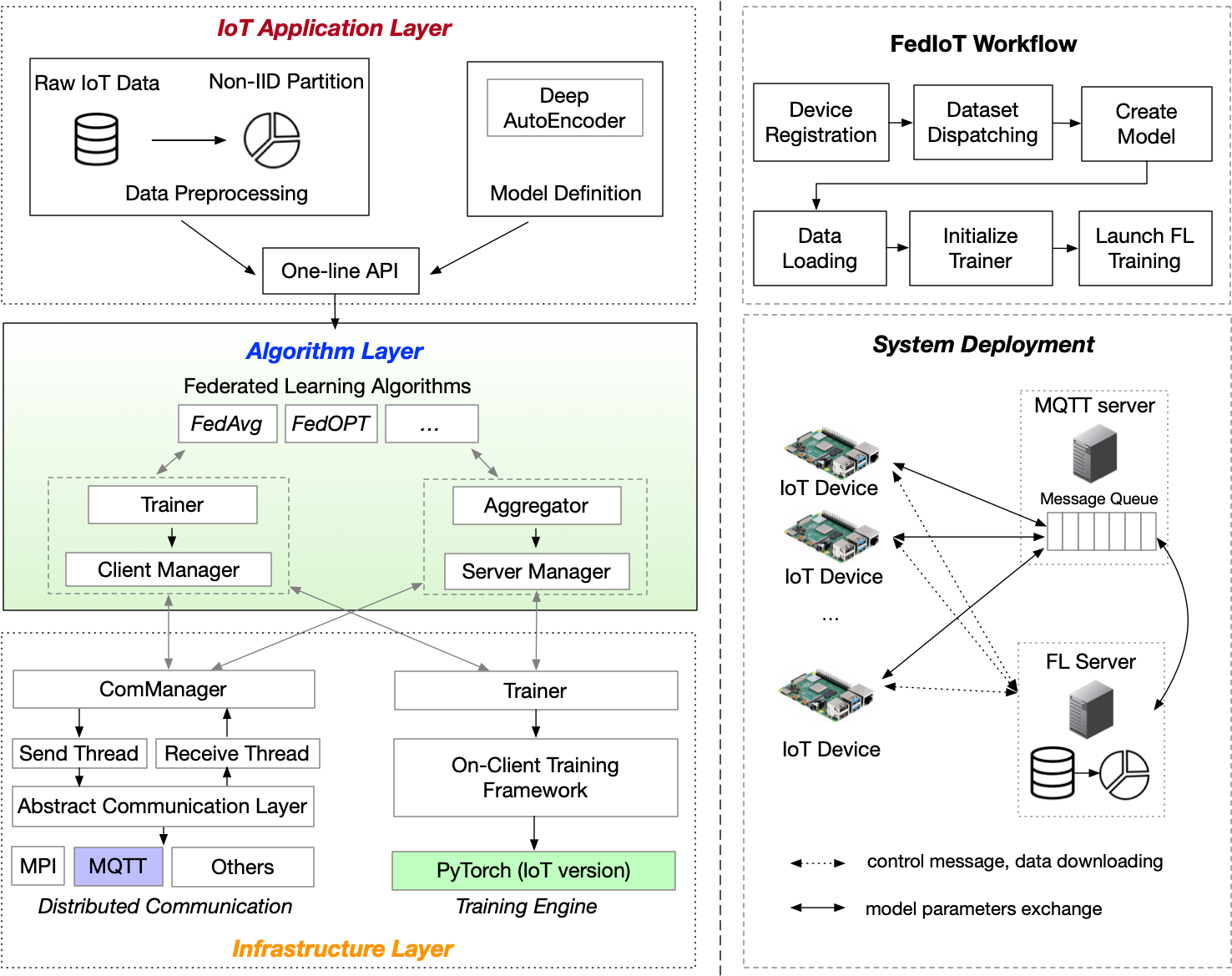}}
\caption{Overview of \texttt{FedIoT} Platform}
\label{fediot-framework}
\end{center}
\vskip -0.2in
\vspace{-0.1in}
\end{figure*}

However, as the traffic volume of IoT-based DDoS attacks reaches unprecedented levels \cite{7842850}, the efficacy of these works is unclear, mainly when the attacks spread to large-scale types and devices but training on the limited data from small-scale devices cannot obtain high accuracy. More significantly, our research community lacks an open, generic, and flexible FL-enabled IoT platform for advanced researches. Existing works only run simulations rather than perform experiments in real IoT platforms, or their specialized system design is not generalized enough for future research. In addition, given that IoT devices are resource-constrained, system performance analysis is also an essential step. Unfornantitely, none of these works provides such analysis towards practical deployment.

To further push forward the research in FL-based IoT cybersecurity, we build \texttt{FedIoT} platform with a simple but effective design philosophy that lays the foundation for future scientific research. The overall design spans dataset, model, algorithm, and system design. More specifically, we propose a federated learning algorithmic framework, \texttt{FedDetect} which utilizes adaptive optimizer (e.g., Adam) and cross-round learning rate scheduler, rather than naive FedAvg \cite{DBLP:journals/corr/McMahanMRA16} for local training.
In order to verify the effectiveness of FL for IoT security, we design a novel method to synthesize the testset from public dataset for FL-based IoT cybersecurity research. Its design aims to evaluate whether the global model obtained through FL training can recognize more attack types and has higher detection performance (the evaluation and test dataset cover all attack types on the entire IoT network). In addition, we build \texttt{FedIoT} platform for realistic IoT devices with a high degree of modularization and flexible APIs. As such, we can add new data, models, and algorithms with only lightweight modification. Most importantly, \texttt{FedIoT} supports both IoT edge training (e.g., Raspberry PI) and CPU/GPU-based distributed training with the support of MQTT and MPI communication backend, respectively.

We evaluate \texttt{FedIoT} platform and \texttt{FedDetect} algorithm on N-BaIoT \cite{Meidan} dataset, and also analyze the system performance comprehensively (i.e., computational speed, communication cost, and memory cost). 
Our results demonstrate the efficacy of federated learning in detecting a large range of attack types. Our system efficiency analysis shows that the training time memory cost occupies only a small fraction of the entire host memory of the IoT platform (Raspberry Pi), and the end-to-end training time is feasible (less than 1 hour).

\section{Algorithm and System Design}

\subsection{Overview}

Federated learning (FL)-based IoT cybersecurity aims to detect network intrusion in IoT devices without centralizing a large amount of high frequent edge data. In this work, to analyze both algorithmic and system performance in real IoT devices, we build \texttt{FedIoT} platform with simple but effective design philosophy, as illustrated in  Figure~\ref{fediot-framework}. The entire software architecture consists of three layers: the application layer, the algorithm layer, and the infrastructure layer. We make each layer and module perform its duty and have a high degree of modularization. In the application layer, \texttt{FedIoT} provides a one-line API (taking the Autoencoder model and data loader as inputs) to launch the federated training on IoT devices in a distributed computing manner. At the algorithm layer, \texttt{FedIoT} supports \texttt{FedDetect} and other algorithmic baselines such as FedAvg \cite{McMahan2017CommunicationEfficientLO}, FedOPT \cite{reddi2020adaptive}; 
at the infrastructure layer, \texttt{FedIoT} aims to support lightweight communication APIs with MQTT \cite{hunkeler2008mqtt} (i.e., Message Queuing Telemetry Transport, a standard for IoT messaging), and IoT platform-customized PyTorch library \cite{paszke2019pytorch} that can support on-device training for CPU or GPU-enabled IoT edge devices, such as Raspberry Pi and NVIDIA Jetson Nano.

We organize the remaining subsections as follows.
Section~\ref{sec:deplotment} depicts an IoT
scenario and threaten model with privacy requirements.
Section~\ref{sec:data} introduces how we synthesize a dataset to understand a larger attack type based on the N-BaIoT dataset.
Section~\ref{sec:model}, Section~\ref{sec:algorithm} and Section~\ref{sec:system} describe the AutoEncoder model, \texttt{FedDetect} algorithm, and the system design, respectively. 


\subsection{Threaten Model and Deployment}
\label{sec:deplotment}
\begin{figure}[h]
\begin{center}
\centerline{\includegraphics[width=0.6\columnwidth]{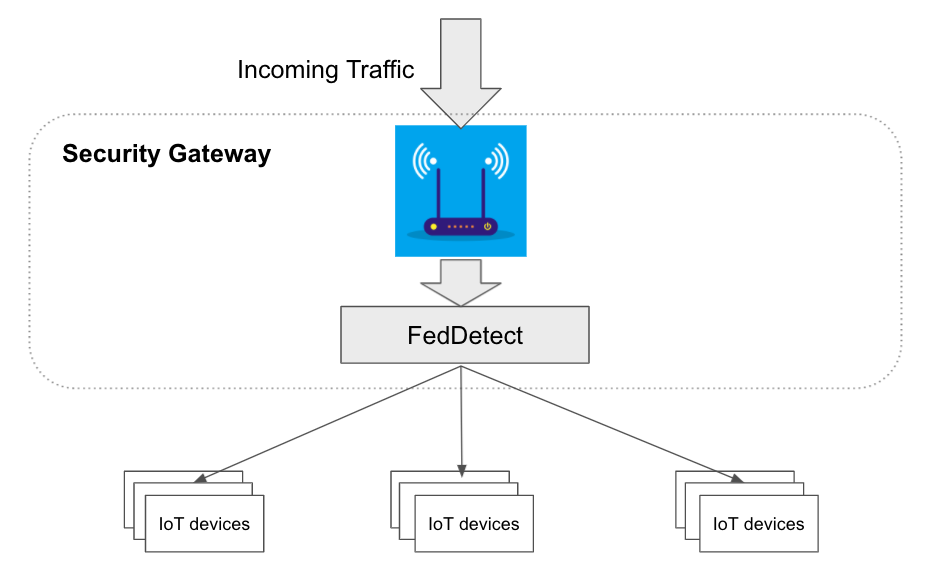}}
\caption{\texttt{FedIoT} deployment}
\label{fediot-deployment}
\end{center}
\vskip -0.35in
\end{figure}

The primary goal for the \texttt{FedDetect} is to prevent the anomaly traffic from the malware, such as the DDoS attack, passing into the IoT devices. As a result, we design to deploy the FedDetect on the router, which makes the router as a security gateway to monitor IoT devices and perform anomaly detection to protect the cybersecurity. Figure~\ref{fediot-deployment} presents the deployment diagram. The incoming traffic will be filtered first by the security gateway before fed into the IoT devices. In this paper, we proposed the following assumptions for the discussion: \textit{1. The IoT device may be vulnerable but not initially be compromised. 2. The Internet gateway (i.e. router) is not compromised}.

\subsection{Dataset and Preprocessing}
\label{sec:data}

We introduce a widely used public IoT dataset for anomaly detection and then synthesize a dataset for realistic evaluation on the FL-based method.

N-BaIoT dataset \cite{Meidan} is a widely used public dataset for research on anomaly detection of IoT data. N-BaIoT captures network traffic flow from 9 commercial IoT devices authentically attacked by network intrusions. Each IoT device has 2 subsets: one benign set containing normal network flow data only, and one attack data subset consisting of two common malware attacks, Mirai and BASHLITE, which each contain five different kinds of attack types.

In order to verify the effectiveness of FL for IoT security, different from previous works  \cite{diot,fldef,jiot,9146846,rey2021federated}, we hope to learn a detection model from benign data widely distributed in different types of devices that can identify a larger range of attacks. Specifically, we hope that the data design meets three characteristics: \textit{1. It contains training data of multiple device types (benign data); 2. Each device has no full set of all attack types; 3. The evaluation and test dataset should cover all attack types on the entire IoT network}. These requirements are based on several real-world assumptions:
\begin{itemize}
    \item From the perspective of benign data, features of the network data flow among different types of devices are inconsistent. For example, a surveillance camera will record in real-time (24x7 hours), while the data generated by a doorbell is intermittent.
    \item The detection model should have the ability to identify multiple attack types, but the types of attacks encountered by a single device are likely to be only part of the full set. Only by learning with the feature of all attack types in the entire IoT network, the detection model on a single device can have the ability to detect unknown attacks in a wide range.
    \item Because of privacy (e.g., camera video) and extremely high communication/storage overhead (e.g., high-frequency data from time-series sensors), it is infeasible to centralize data on massive devices.
\end{itemize}

Therefore, we use N-BaIoT to synthesize the testset for FL-based IoT cybersecurity research. Our synthesized testset is generated by the following rules: 
1. For each device, we assign 5000 samples of selected benign data as the training data and another 3000 samples of benign data as the evaluation dataset (i.e., calculating the anomaly threshold, see Section \ref{sec:model}); 2. We enforce the global test dataset to compose all devices’ benign data and all types of attack data. More specifically, for each device, we randomly select 500 malicious data samples from each type of the attack, and the same amount of benign data samples to synthesis the testset, making the testset perfectly balanced for binary classification.

\subsection{Anomaly Detection with Deep Autoencoder}
\label{sec:model}

We apply a Deep Autoencoder \cite{rumelhart1985learning} as the model for anomaly detection. Deep Autoencoder is simple but effective and does not lose the generality to evaluate FL algorithms and our \texttt{FedIoT} platform. Other advanced models (e.g., Variational Autoencoder, or attention-based CNN-LSTM \cite{9146846}) can also be applied into our framework without additional engineering efforts.

\paragraph{Model Definition} Deep Autoencoder focuses on the reconstruction of the input data in an unsupervised learning manner. Essentially, Autoencoder splits the neural network into two segments, the encoder $f_{\theta_{e}}$ and the decoder $f_{\theta_{d}}$. Encoder $f_{\theta_{e}}$ compresses the input $\mathbf{x}$ to a latent space $\mathbf{z}$. Decoder $f_{\theta_{d}}$ then attempts to restore the original image after some generalized non-linear transformation. Mathematically, the loss function can be written as 
$\mathcal{L}\left(\mathbf{x}, \mathbf{x}^{\prime}\right)=\left\|\mathbf{x}-\mathbf{x}^{\prime}\right\|^{2} = \left\|\mathbf{x}- f_{\theta_{d}}(\mathbf{z}) \right\|^{2} = \left\|\mathbf{x}- f_{\theta_{d}}(f_{\theta_{e}}(\mathbf{x})) \right\|^{2}$. This loss is also called \textit{reconstruction error} calculated by mean square error $MSE = \frac{1}{d}\sum_{i = 1}^{d}\left ( x_i - \hat{x_i} \right)^2$, where $d$ is the dimension of the input. 
In essence, this loss function aims to encode the input to a latent representation $\mathbf{z}$ such that it can be regenerated by the decoder. To minimize the loss, common deep learning optimizers such as Adam \cite{kingma2014adam} can be applied.

\begin{small}
\begin{equation}
tr = \overline{MSE} + \alpha*\sigma(MSE)
\label{eq:threshold}
\end{equation}
\vskip -0.2in
\end{small}

\paragraph{Anomaly Detection} 
In the application of anomaly detection, we train a deep Autoencoder with benign IoT traffic data to learn IoT devices' normal behavior, so that our Autoencoder could successfully extract and reconstruct features on benign samples but fails to do so on abnormal samples with unseen features. During detection phase, one input data sample that achieves a reconstruction error above a threshold will be detected as an abnormal data sample. In detail, after training Autoencoder on benign training dataset, we first calculate the reconstruction error ($MSE$) for each data sample from benign evaluation dataset, and then obtain the threshold by Equation~\ref{eq:threshold}, which computes the sum of the mean of $MSE$ plus standard deviation of $MSE$ over all evaluation samples. The value of threshold should be as large as possible to suppress the majority of benign samples while preventing abnormal samples from being classified into benign samples. Extensive experiments show that the overall performance is the best when $\alpha$ equals to 3.

\subsection{FedDetect}
\label{sec:algorithm}

We propose a federated learning algorithmic framework, \texttt{FedDetect}, for anomaly detection in distributed IoT devices.
Distinguished from existing works on FL-based IoT anomaly detection, \texttt{FedDetect} utilizes adaptive optimizer (e.g., Adam) and cross-round learning rate scheduler, rather than naive FedAvg \cite{DBLP:journals/corr/McMahanMRA16} for local training. \texttt{FedDetect} is summarized as Algorithm \ref{alg:fedavg}.

\newcommand{\adjustlr}{\colorbox{blue!18}{Adjust cross-round learning rate (cosine scheduler)}}
\newcommand{\clienttraining}{\colorbox{green!18}{Local Training with Adam}}


\begin{algorithm}[h!]
   \caption{\texttt{FedDetect}}
   \label{alg:fedavg}
\begin{algorithmic}[1]
   \STATE {\bfseries Initialization} $w_0$

   \FOR{ round $t$= 0, 1, ...} 
       \STATE 
       \adjustlr
       \FOR{client $i=0$ {\bfseries to} $K-1$}
       \STATE ${w_{t+1}}^i\leftarrow$ 
       \clienttraining
       \STATE Upload ${w_{t+1}}^i$ to Server
       \ENDFOR
   \STATE ${w_{t+1}} = \frac{1}{K}\sum_{i=0}^{K-1} {w_{t+1}}^i$ 
   \STATE clients receive new model ${w_{t+1}}$
   \ENDFOR
   \STATE {\bfseries Globalized Threshold Algorithm}
   \STATE $MSE_{Global} = [MSE^0, ..., MSE^{K-1}]$ 
   \STATE $tr_{Global} = \overline{MSE_{Global}} + \alpha*\sigma(MSE_{Global})$
\end{algorithmic}
\end{algorithm}

\paragraph{Local Adaptivity and Cross-round Learning Rate Scheduler} The choice of Adam for local training and cross-round learning rate scheduler is based on our experimental observation. We empirically find that local Adam beats naive local SGD or SGD with momentum when applying a cross-round learning rate scheduler (e.g., cosine scheduler). The rationality of local adaptivity has been intensively verified its efficacy in CV and NLP tasks \cite{Wang2021LocalAI,reddi2020adaptive}.

\paragraph{Global Threshold} After achieving the global model via federated training, we propose a algorithmic module to calculate the anomaly threshold for each device, \textit{Global Threshold}.
More specially, in the Global Threshold algorithm, each device runs the global model locally to get the $MSE$ sequence and then synchronizes it to the server, and the server uses MSE sequences from all devices to generate a unified global threshold for detection in each device. 

Global Threshold algorithm objectively expresses the detection performance of the FL-trained global model on the entire IoT networks (i.e., detecting a larger range of attack types as introduced in Section \ref{sec:data}). Experimental results demonstrate the efficacy of FL in diverse real-world scenarios (Section \ref{sec:exp_fl}).

\subsection{System Design}
\label{sec:system}
As illustrated in Figure \ref{fediot-framework}, the entire software architecture of \texttt{FedIoT} platform consists of three layers: the application layer, the algorithm layer, and the infrastructure layer. \texttt{FedIoT} aims to provide an open, generic and flexible FL-enabled IoT platform for future research exploration. 

In the application layer, APIs for data preprocessing described in Section \ref{sec:data} are provided. Combined with the model definition, we could launch the FL on IoT platforms using a one-line API. The figure also shows the workflow, which is a standard engineering practice that we suggest. 

In the algorithm layer, beside the regular \texttt{FedAvg} and \texttt{FedOPT} FL aggregation functions, \texttt{FedIoT} supports customized local trainer, dataloader and FL aggregation functions. The FL client (IoT device) is maintained as an instance by the \texttt{ClientManager}, and the FL server is managed by \texttt{ServerManager}. They can exchange information with flexible APIs supported by the underlying abstract communication layer. By following this client-oriented programming design pattern, users can define their own client/server behavior when participating in training or coordination in the FL algorithm, so that \texttt{FedIoT} allows users to focus on algorithmic implementations instead of considering the low-level communication mechanism. 

In the infrastructure layer, \texttt{FedIoT} supports AI-enabled IoT edge devices and the lightweight communication APIs for distributed system. With such testbeds built upon real-world hardware platforms, researchers can evaluate realistic system performance, such as training time, communication, and computation cost. Moreover, for \texttt{FedIoT}, researchers only need to program with Python to customize their research experiments without the need to learn new system frameworks or programming languages (e.g., Java, C/C++). 
MQTT \cite{hunkeler2008mqtt} is supported for IoT setting. As such, we deploy two servers: the MQTT server for messaging passing and the FL server for federated aggregation. The model weight exchange is by MQTT messaging server, while control messages such as device registration and dataset distribution are transmitted through the FL server.

With these design, researchers can add new data, models, and algorithms with only lightweight modification.


\section{Experiments}
\label{sec:experiments}
We evaluated \texttt{FedIoT} platform on two aspects: algorithmic performance in global model; a comprehensive analysis of the system efficiency, including computational speed, communication cost, and memory cost.

\subsection{Setup}


\paragraph{Implementation} We implemented two computing paradigms of \texttt{FedIoT} platforms: (1) IoT edge training, and (2) CPU/GPU-based distributed training. For the IoT edge training, we choose 9 Raspberry Pi 4B as client devices and a GPU server that integrates both the FL server and the MQTT service. The local training will be implemented on the Raspberry Pi and the weight integration will be implemented on the GPU server.

\paragraph{Dataset and Model Definitions} We evaluated \texttt{FedIoT} platform using the dataset described in Section \ref{sec:data}. For the Autoencoder, we set the input with a dimension of 115, the same as the features of data. The encoder network has four hidden layers, as the dimension decreasing rate equals 75\%, 50\%, 33\%, and 25\%. The decoder has the same layer design as the encoder, but with an increasing sequence. The number of parameters for the Autoencoder is equal to $36628$. 

For the training dataset, we implement the min-max normalization, i.e. $x = \frac{x - x^{min}}{x^{max}-x^{min}}$, where $x^{min}$ and $x^{max}$ are the global minimum and maximum values over training samples of all devices. As for the evaluation and testing dataset, we use the same minimum and maximum values obtained from the training dataset to scale both benign and attack samples, which will make the testing data as the same distribution as benign training samples.
This scaling will distort the original distribution for the malicious data, which decreases the restoration performance by the Autoencoder and assist the performance for the binary classification.

\paragraph{Hyper-parameters} We searched for the learning rate on a range of \{1, 0.1, 0.01, 0.001, 0.0001, 0.00001\}, input batch size on a range of \{8, 16, 32, 64, 128\}, local training epoch on a range of \{1, 50, 100, 120, 150, 200\}, total training round on a range of \{10, 30, 50, 100, 120, 150\} and tested both $tanh$ and $sigmoid$ activation functions in Autoencoder. After hyper-parameter searching, we fixed the following hyper-parameters: the batch size for input is 64, the local training epoch in FL is 120, total training round is 30, and the activation function inside the Autoencoder is set as $tanh$ function. More hyper-parameters can be found in our source code.

\subsection{Baselines}

To evaluate our proposed algorithm in Section \ref{sec:algorithm} comprehensively, we design the following scenarios for the direct comparison:

\begin{itemize}
  \item \textit{CL-Single}: Each device trains a detection model with its own local dataset. This baseline can obtain the model performance when training a local model without the help of federated learning.
  \item \textit{CL-Combined}: A detection model is trained with the merged data from nine devices. The result of this baseline serves as the upper bound performance for the centralized training. It may perform the best because it gathers all data samples from the entire IoT network.
  \item \textit{FL-FedDect}: A detection model is trained via federated learning from nine devices with the \texttt{FedDect} algorithm. It may give similar performance as CL-Combined.
\end{itemize}

\subsection{Metrics}
\label{sec:metrics}
Following the existing works, we use three metrics to evaluate the detection performance: accuracy (ACC), true positive rate (TPR), false positive rate (FPR), and true negative rate (TNR). The formulas and confusion matrix are shown below.
\begin{multicols}{2}
    \begin{itemize}
    \addtolength\itemsep{-0.5mm}
        \item $ACC = \frac{TP + TN}{TP + TN + FP + FN}$
        \item $FPR = \frac{FP}{TN + FP}$
        \item $TPR = \frac{TP}{TP + FN}$
        \item $TNR = \frac{TN}{TN + FP}$
    \end{itemize}
\end{multicols}
\begin{table}[h!]
\label{tb-confusion}
\small
\centering
\begin{tabular}{@{}lll@{}}
\toprule 
                                      & Predicted Benign & Predicted Malicious \\ \midrule
\multicolumn{1}{l|}{Actual Benign}    & True Negative (TN)                 & False Positive (FP)\\
\multicolumn{1}{l|}{Actual Malicious} & False Negative (FN)                & True Positive (TP) \\ 
\bottomrule 
\end{tabular}
\end{table}

\subsection{Results of Learning Performance}
\label{sec:exp_fl}

\paragraph{Evaluation using the global threshold}  We first evaluated the performance of \texttt{FedDetect} algorithm using the global threshold and compared its performance with three baselines. For the baseline CL-Single, we reported the average value of the nine devices' model performances. The detailed results for the three scenarios are shown in the Table~\ref{tb:results}.

\begin{table}[t]
\small
\centering
\caption{
Performance of anomaly detection under both centralized training and federated training
}
\begin{tabular}{|c|c|c|c|c|} \hline 
 & Acc & FPR & TPR & TNR \\ \hline \hline
\textit{CL-Single} & 73.82\% & 37.50\% & 86.56\% & 62.50\%\\ \hline
\textit{CL-Combined} & 98.64\% & 2.71\% & 99.99\% & 97.29\% \\ \hline
\textit{FL-FedDect} & 98.27\% & 3.45\% & 99.99\% & 96.55\%  \\ \hline \hline
\end{tabular}
\label{tb:results}
\end{table}

\paragraph{Understanding the result} 
For the detection model, the more benign data the model can train on, the better performance it should have. Under the global evaluation, the baseline CL-Combined trains on all benign data. Consequently, it is the best performance among all models. Within the same amount of training samples, FL with distributed training achieves nearly the same performance compared to upper bound of centralized training, which demonstrates the efficiency of \texttt{FedDetect}.  The \texttt{FedDetect} algorithm has much better performance than CL-Single, because the \texttt{FedDetect} trains on more data among all devices collaboratively, thus it can capture more features. The FPR and TNR performance of FL is little worse than CL, because during ML optimization, the direction of gradient descent is shifted after the aggregation by averaging, leading to sub-optimal minimum, which may not be suitable for the local evaluation targets.

\subsection{Analysis of System Efficiency}
\label{sec:system_analysis}

For the second part of the experiment, we evaluated the system performance of \texttt{FedIoT} with globalized threshold on the Raspberry Pi within N-BaIoT dataset. 

\begin{table}[t]
\small
\centering
\caption{
CPU/GPU Training v.s. IoT Edge Training
}
\begin{tabular}{|c|c|c|c|c|} \hline 
 & Acc & FPR & TPR & TNR \\ \hline \hline
\textit{Simulation} & 98.27\% & 3.45\% & 99.99\% & 96.55\% \\ \hline
\textit{Raspberry Pi} & 97.47\% & 4.78\% & 99.99\% & 95.22\%  \\ \hline \hline
\end{tabular}
\label{tb-CvsR}
\end{table}

We first verified that \texttt{FedIoT} on the real IoT device could achieve the same results as CPU/GPU distributed training. From Table \ref{tb-CvsR}, we could see that the results from the Raspberry Pi are nearly the same as the results from CPU/GPU simulation. The slight difference is due to different random initialization (i.e., different runs in different platforms).

\begin{figure}[h!]
\centering
\subfigure[Memory Percentage]{
\begin{minipage}[t]{0.5\linewidth}
\centering
\includegraphics[width=0.6\columnwidth]{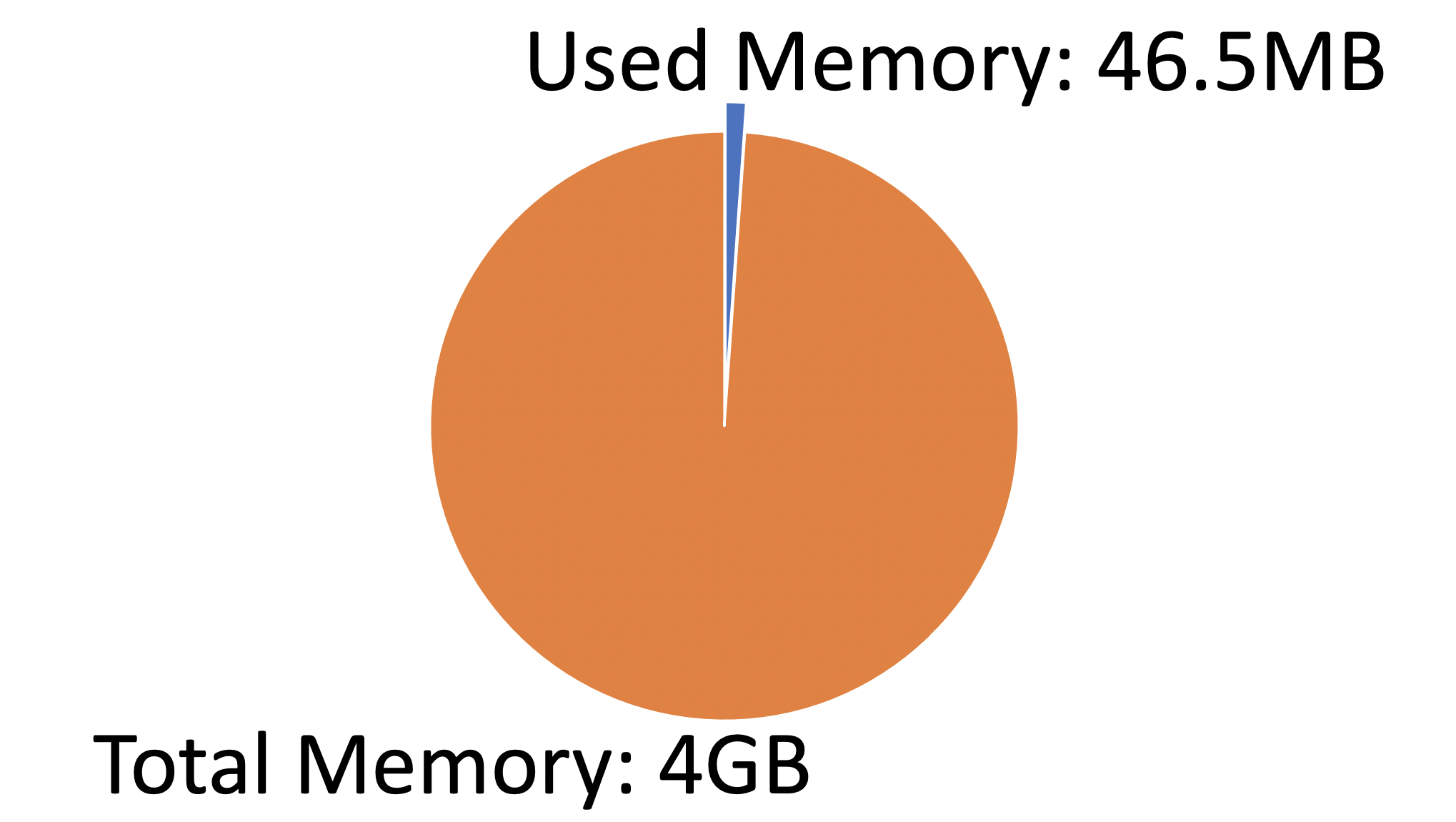}
\end{minipage}%
}%
\subfigure[Running Time Per Round]{
\begin{minipage}[t]{0.5\linewidth}
\centering
\includegraphics[width=0.6\columnwidth]{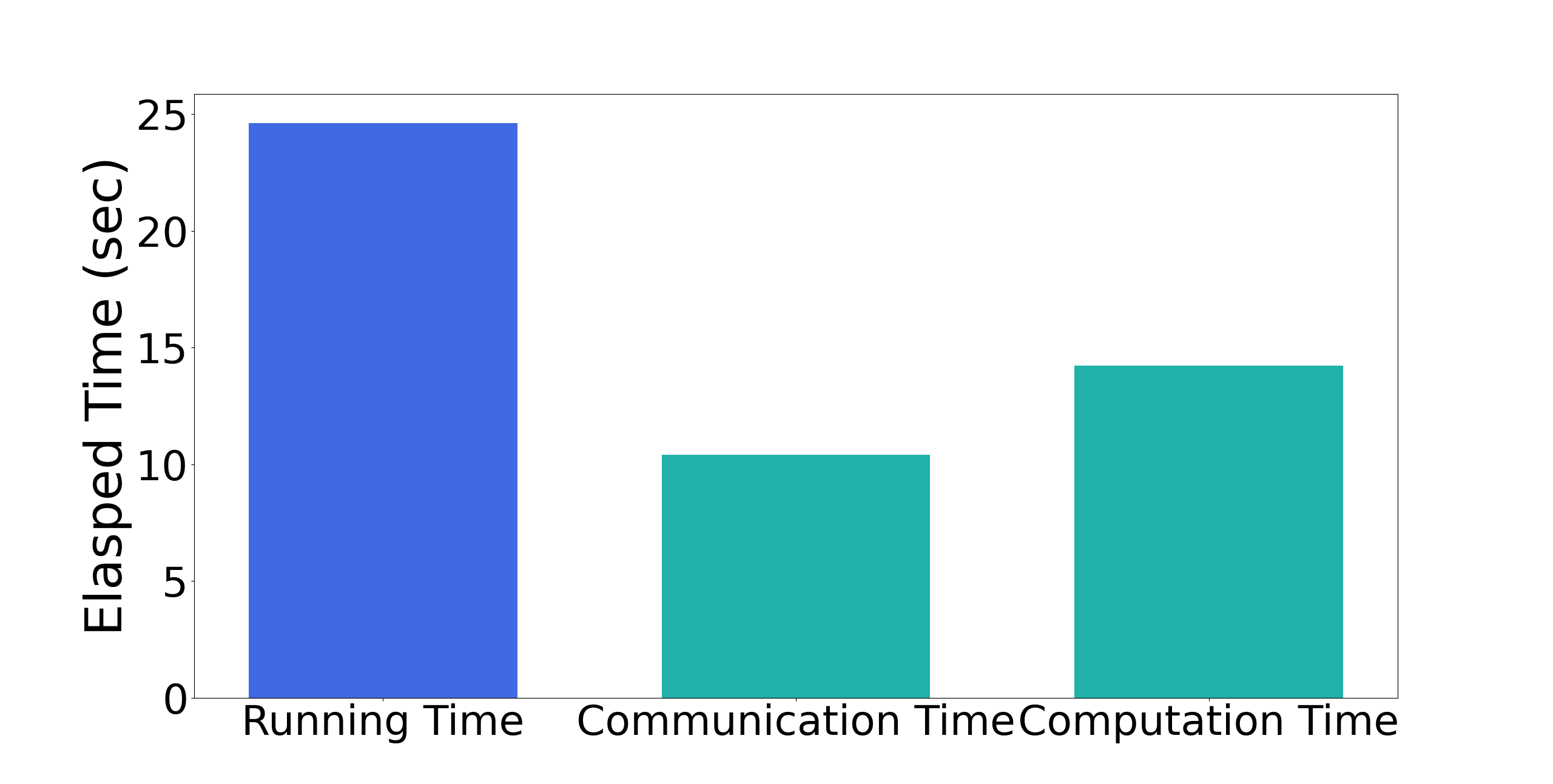}
\end{minipage}%
}%
\centering
\caption{Properties of Experiments on Raspberry Pi}
\label{fg-memory_time}
\end{figure}

We further tested the system efficiency on the Raspberry Pi platform. From Figure \ref{fg-memory_time}, we can see that the training time memory cost occupies only a small fraction of the entire host memory of Raspberry Pi (only 4G host memory). The training time per round is less than 1 minute. To understand the system cost more comprehensively, we analyzed the breakdown of the end-to-end training time when the bandwidth is 7.65MB/s (a reasonable bandwidth in 4G/5G wireless communication), and results are shown in Table \ref{tb-rasp}. Overall, the end-to-end training time is an acceptable training time (less than 1 hour) for practical applications. We can also find that the ratio of communication almost costs half of the end-to-end training time, indicating that the communication compression technique \cite{lin2017deep,tang2018communication} is essential to improve the system performance in the IoT setting. 

\begin{table}[t]
\caption{Breakdown of the End-to-end Training Time}
\label{tb-rasp}
\begin{center}
\small
\begin{tabular}{p{150pt}p{60pt}}
\toprule
Type & Value \\ \midrule
end-to-end time        & 2547 seconds\\ \midrule
uplink latency     & 0.167 seconds\\ \midrule
communication time ratio        & 42.2 $\%$ \\ \midrule
computation time ratio        & 57.8 $\%$\\ \midrule
bandwidth              & 7.65 MB/s\\ 
\bottomrule
\end{tabular}
\end{center}

\footnotesize
*Note: the communication time is measured by computing the interval between the timing when Raspberry Pi uploads local model to the server and the timing that Raspberry Pi receives the global model from the server. The experiment is implemented under the WiFi condition.

\end{table}

\section{Related Works}
\label{sec:related}
Our work is related to the application of federated learning in IoT cybersecurity.
DIoT \cite{diot} is the first system to employ a federated learning approach to anomaly-detection-based intrusion detection in IoT devices. IoTDefender \cite{fldef} is another similar framework but  obtains a personalized model by fine-tuning the global model trained with federated learning.
\cite{jiot} evaluates FL-based anomaly detection framework with learning tasks such as aggressive driving detection and human activity recognition.
\cite{9146846} further proposed an attention-based CNN-LSTM model to detect anomalies in an FL manner, and reduced the communication cost by using Top-$k$ gradient compression. Recently, \cite{rey2021federated} even evaluates the impact of malicious clients under the setting of FL-based anomaly detection. Compared to these existing works, our FedIoT platform is the first work that analyzes both algorithmic and system performance in a real IoT platform.

\section{Conclusion}
\label{sec:conclusion}
In this paper, to further push forward the research in FL-based IoT cybersecurity, we build \texttt{FedIoT} platform with a simple but effective design philosophy. We apply Deep Autoencoder \cite{rumelhart1985learning} as the model for anomaly detection to evaluate FL algorithms and our \texttt{FedIoT} platform. Moreover, we propose \texttt{FedDetect}, a federated learning algorithmic framework that utilizes adaptive optimizer and cross-round learning rate scheduler, rather than naive FedAvg \cite{DBLP:journals/corr/McMahanMRA16} for local training. \texttt{FedIoT} supports both IoT edge training and CPU/GPU-based distributed training with the support of MQTT and MPI communication backend, respectively. We evaluate \texttt{FedIoT} platform and \texttt{FedDetect} algorithm with global detection model, and also analyze the system performance comprehensively. Our results demonstrate the efficacy of federated learning in detecting a large range of attack types, and the system efficiency analysis shows that both end-to-end training time and memory cost is affordable and promising for resource-constrained IoT devices.

\bibliography{iclr2022_conference}
\bibliographystyle{unsrtnat}
\end{document}